\documentclass[10pt,twocolumn,letterpaper]{article}

\usepackage{cvpr}              

\usepackage{placeins}

\definecolor{cvprblue}{rgb}{0.21,0.49,0.74}
\usepackage{array}
\usepackage{booktabs}
\usepackage{float}
\usepackage[pagebackref,breaklinks,colorlinks,allcolors=cvprblue]{hyperref}

\def\paperID{*****}
\def\confName{CVPR}
\def\confYear{2026}

\title{AutoMine Solution for AV2 2026 Scenario Mining Challenge}

\author{
Songliang~Cao$^{1,2}$~\thanks{This work was done during internship at Xiaomi.}
\quad
Jiele~Zhao$^{1}$~\footnotemark[1] 
\quad
Yuru~Wang$^{1}$
\quad
Hao~Li$^{1}$
\quad
Daqi~Liu$^{1}$ 
\quad
Zehan~Zhang$^{1}$~\thanks{Project leader.}
\\
Fangzhen~Li$^{1}$~\footnotemark[2]
\quad
Yu Wang
\quad
Yue Zhang
\quad
Bing~Wang$^{1}$
\quad
Guang~Chen$^{1}$ 
\quad
Hao~Lu$^{2}$
\quad
Hangjun~Ye$^{1}$
\\[2mm]
$^{1}$Xiaomi~EV\quad
$^{2}$Huazhong University of Science and Technology
}

\begin{document}
\maketitle

\begin{abstract}
With the development of autonomous driving systems, mining high-value, safety-critical, and planning-relevant scenarios from large-scale driving logs has become essential for data-driven evaluation. In this paper, we propose \textbf{AutoMine}, a robust self-refining scenario mining method based on LLMs and VLMs. AutoMine uses semantics-preserving prompt augmentation to reduce LLM prompt sensitivity, combines robust trajectory atomic functions with VLM-based functions to handle perception noise and open-world visual cues, and refines generated code through execution feedback from real logs. In the Argoverse 2 Scenario Mining Competition at CVPR 2026, AutoMine achieves a HOTA-Temporal score of \textbf{36.38} and a Timestamp BA score of \textbf{77.21}.
\end{abstract}

\section{Introduction}

Autonomous driving datasets contain massive sensor logs, while rare and safety-critical events remain sparse. Scenario mining enables targeted evaluation by retrieving logs, timestamps, and 3D actors that match a natural language description.

This task is challenging because query wording is precise, predicted tracks are noisy, and some scenarios require visual evidence beyond 3D trajectories. For example, \textit{passing} and \textit{overtaking} may imply different conditions, while tracks may contain missing detections, heading noise, fragmentation, and ID switches.

The rapid development of LLMs and VLMs brings new possibilities for scenario mining. RefProg~\cite{davidson2025refav} shows that LLMs can translate natural language descriptions into composable atomic function calls for interpretable trajectory-based mining. However, manually specified atomic functions are hard to scale to open-world visual concepts. In addition, one-shot generated code can fail when the LLM misunderstands the prompt, selects the wrong category, reverses a relation, or imposes incorrect constraints; this is consistent with prior findings on LLM sensitivity to meaning-preserving prompt design choices~\cite{sclar2024quantifying}.

We propose \textbf{AutoMine}, a multimodal scenario mining framework that strengthens LLM-generated programs with semantic-preserving prompt augmentation, robust atomic functions, VLM-based visual functions, perception post-processing, and execution-driven self-refinement. AutoMine executes generated code on real logs, summarizes the observed outputs, and uses the feedback to repair systematic errors, improving robustness to language ambiguity and perception noise.
\begin{figure*}[t]
    \centering
    \includegraphics[width=\textwidth]{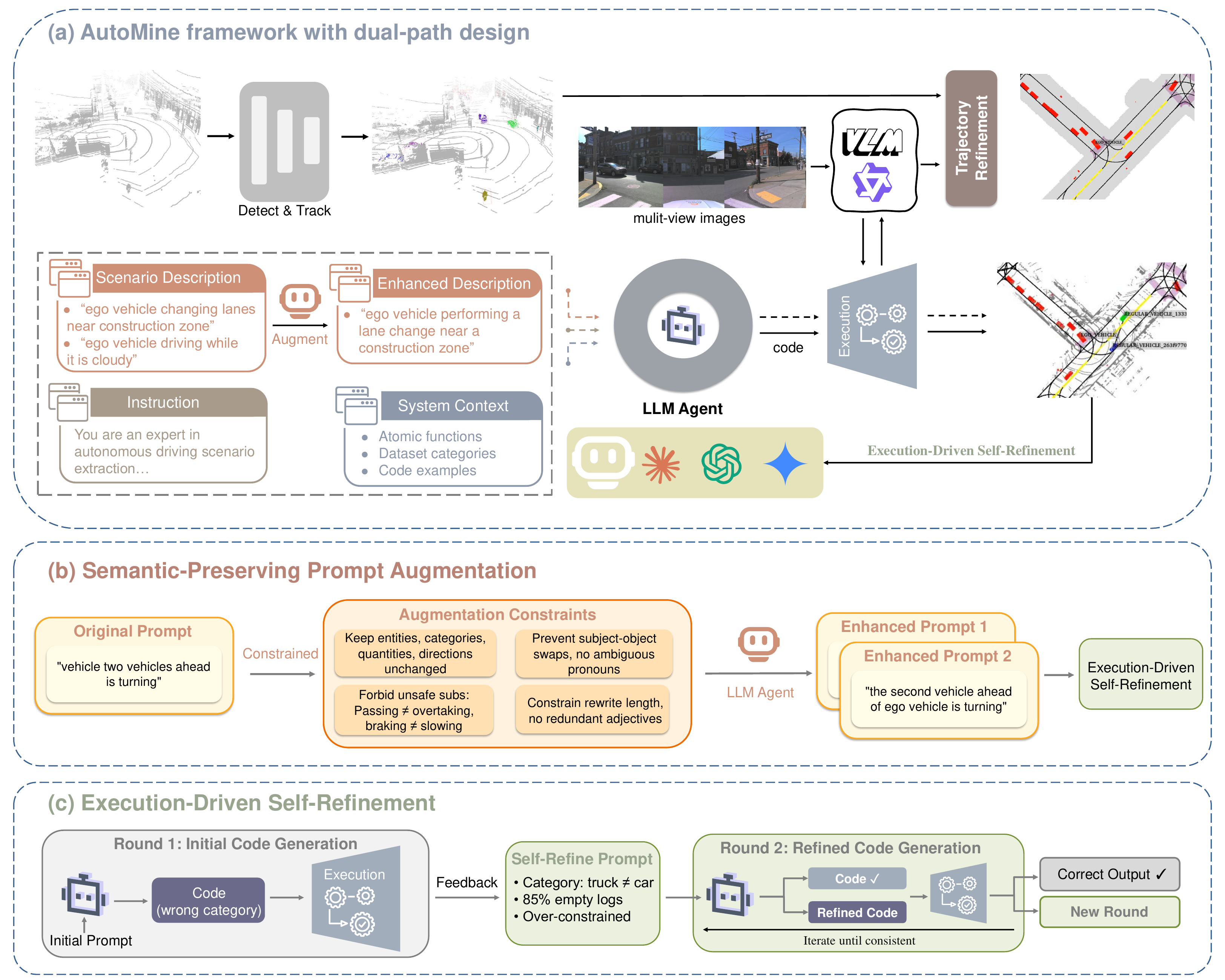}
    \caption{ (a): Overview of the AutoMine framework with dual-path design (perception + language). (b): Semantic-preserving prompt augmentation. (c): Execution-driven self-refinement loop.
}
    \label{fig:overview}
\end{figure*}
\section{Method}

\subsection{Overview}

Given a natural language description, driving logs, and sensor data, AutoMine outputs referred actors and valid timestamps. As shown in Fig.\ref{fig:overview}, AutoMine first refines perception tracks, then combines semantic-preserving prompt augmentation, LLM-generated atomic functions, multimodal execution over tracks, maps, and images, and execution-feedback-based code refinement. We describe each module in the following sections.

\subsection{Trajectory Refinement}

AutoMine uses the detection and tracking results from Le3DE2E~\cite{wang2023le3de2e} as initial trajectory inputs. Although these results provide strong 3D tracks, we observe ID switches, fragmented trajectories, missed detections, false positives, and duplicated boxes, which directly affect temporal localization and referred-actor selection.

Inspired by Immortal Tracker~\cite{wang2021immortal}, we keep short unmatched tracklets alive instead of terminating them immediately, and reconnect compatible fragments using spatial, temporal, category, and size consistency. We also apply backward tracking to recover early segments that are easier to associate from later frames. In addition, we use the grounding capability of Qwen3.5-27B~\cite{qwen3.5} to verify projected boxes in camera views and remove additional false positives. This trajectory refinement improves track continuity and provides more reliable inputs for downstream atomic functions.

\subsection{Semantic-Preserving Prompt Augmentation}

As mentioned above, prior work shows that LLMs are sensitive to prompt wording and formatting~\cite{sclar2024quantifying}, and we observe the same issue when generating scenario mining code. To reduce this instability, AutoMine augments the natural language scenario descriptions before code generation.

The augmentation is constrained to preserve the original semantics rather than freely paraphrase the query. The rewrite prompt explicitly keeps all entities, categories, quantities, directions, road context, spatial-temporal relations, numerical values, and the referred target unchanged. It also forbids unsafe substitutions such as \textit{passing} $\neq$ \textit{overtaking}, \textit{braking} $\neq$ \textit{slowing}, \textit{changing lanes} $\neq$ \textit{merging}, \textit{stopped} $\neq$ \textit{parked}, and \textit{near} $\neq$ \textit{next to}. To avoid misleading the LLM's judgment of referred objects, related objects, and relation directions, we further enforce constraints that prevent subject-object swaps, demoting the referred actor into a modifier, changing category granularity, or introducing ambiguous pronouns. We also constrain the rewrite length to prevent over-introducing redundant adjectives or extra background.

\subsection{Robust Trajectory-Based Atomic Functions}

AutoMine represents scenario logic as compositions of atomic functions over tracks, ego poses, maps, and time windows. Since predicted tracks are noisy, we refactor motion and relation functions to use temporally aggregated evidence instead of brittle single-frame measurements. Directional and spatial relations are evaluated over multiple valid frames, with relaxed continuity checks and ego-relative geometry.

We also extend the library for planning-centric behaviors not covered by the baseline functions, including U-turns, three-point turns, side parking, special stopping behavior, object interactions, and map-aware road constraints. We use LLM clustering over all descriptions to identify missing function categories, then manually revise and implement the final function set.

\subsection{VLM-Enhanced Atomic Functions}

Some scenarios require visual evidence beyond 3D tracks and maps. AutoMine therefore provides VLM-enhanced atomic functions for fine-grained object type, visual attributes, environment, road surface, zones, pedestrian actions, traffic lights, occlusion, and attached objects. We use Qwen3.5-27B~\cite{qwen3.5} as the underlying vision-language model for all visual reasoning calls.

For visual reasoning about candidate actors, we collect all camera views where the candidate appears and draw the projected 3D box on the original image instead of cropping it, preserving context under projection noise. For environment-level conditions, we use a representative front-view image because each log is short and the environment is usually stable. For road and zone conditions, we stitch candidate-visible frames into a panel annotated with camera names and timestamp indices, and the VLM returns the valid timestamps. Ego-related visual functions use separate prompts because the ego vehicle is not represented by a normal projected object box.

The VLM-enhanced function library includes the functions in Table~\ref{tab:vlm_functions}. These functions allow AutoMine to preserve the advantages of symbolic program execution while extending coverage to open-world visual concepts.

\begin{table}[t]
\centering
\small
\renewcommand{\arraystretch}{1.15}
\setlength{\tabcolsep}{4pt}
\caption{VLM-enhanced atomic functions.}
\label{tab:vlm_functions}
\begin{tabular}{|>{\bfseries\raggedright\arraybackslash}p{0.44\linewidth}|>{\raggedright\arraybackslash}p{0.43\linewidth}|}
\hline
\textbf{Function} & Description \\
\hline
\begin{tabular}[t]{@{}l@{}}\texttt{is\_specific\_}\\\texttt{object\_type()}\end{tabular} & Identifies fine-grained object types. \\
\hline
\texttt{in\_environment()} & Checks weather or environment conditions. \\
\hline
\texttt{has\_attribute()} & Verifies visual attributes or states. \\
\hline
\texttt{on\_road\_type()} & Determines the road type. \\
\hline
\begin{tabular}[t]{@{}l@{}}\texttt{on\_road\_}\\\texttt{surface()}\end{tabular} & Checks road surface conditions. \\
\hline
\texttt{in\_zone()} & Determines whether an actor is in a visual zone. \\
\hline
\begin{tabular}[t]{@{}l@{}}\texttt{pedestrian\_}\\\texttt{action()}\end{tabular} & Recognizes pedestrian actions. \\
\hline
\texttt{object\_color()} & Identifies object color. \\
\hline
\begin{tabular}[t]{@{}l@{}}\texttt{at\_traffic\_}\\\texttt{lights()}\end{tabular} & Checks traffic-light-related conditions. \\
\hline
\texttt{is\_occluded\_by()} & Detects visual occlusion relations. \\
\hline
\begin{tabular}[t]{@{}l@{}}\texttt{has\_attached\_}\\\texttt{object()}\end{tabular} & Checks attached or carried objects. \\
\hline
\end{tabular}
\end{table}

\subsection{Execution-Driven Self-Refinement}

LLM-generated mining code often fails in systematic ways, such as selecting the wrong referred category, reversing relation-function arguments, missing \texttt{reverse\_relationship}, using overly strict thresholds, or misjudging front/back and left/right geometry. Since these errors are hard to identify from code alone, AutoMine refines code with feedback from real execution.

In each round, a code generator produces scenario-mining code, and an executor runs it on up to \texttt{max\_logs} logs with trajectory and VLM atomic functions. AutoMine then summarizes what the code retrieves, including per-track category, temporal coverage, size, ego-object geometry, and cross-log statistics such as referred-category distribution, empty-log ratio, and related-object distribution. These diagnostics expose category confusion, noisy short tracks, over-constrained logic, and relation-direction errors.

The next refinement prompt includes the original description, function library, category definitions, previous code, structured feedback, and the referred category from \texttt{REFERRED\_DICT} as a hard constraint. The LLM keeps the code unchanged if the feedback is consistent; otherwise, it repairs function choices, argument order, reverse relations, thresholds, category filters, or spatial reasoning. This forms an execution-grounded self-refinement loop.

\begin{table*}[t]
\centering
\small
\renewcommand{\arraystretch}{1.1}
\setlength{\tabcolsep}{4pt}
\caption{Ablation results on the Argoverse 2 validation set.}
\label{tab:ablation}
\begin{tabular}{lcccc}
\toprule
\textbf{Configuration} & \textbf{HOTA-Temporal}($\uparrow$) & \textbf{HOTA-Track}($\uparrow$) & \textbf{Timestamp BA}($\uparrow$) & \textbf{Log BA}($\uparrow$) \\
\midrule
\multicolumn{5}{l}{\textbf{Different LLM backends (single-query baseline)}} \\
gpt-5.3-codex-9 & 15.72 & 24.93 & 61.19 & 63.76 \\
Gemini-2.5-Pro & 21.32 & 29.54 & 64.49 & 64.95 \\
Claude-Sonnet-4.6 & 22.14 & 31.05 & 67.17 & 67.97 \\
\midrule
\multicolumn{5}{l}{\textbf{Ablations on Claude-Sonnet-4.6}} \\
+ Trajectory Refinement & 24.04 & 35.96 & 67.51 & 71.59 \\
+ Atomic Function Optimization (Robust + VLM) & 31.46 & 41.86 & 74.61 & 76.98 \\
+ Execution-Driven Self-Refinement & 33.96 & 44.95 & 75.98 & \textbf{80.04} \\
+ Semantic-Preserving Prompt Augmentation & \textbf{34.99} & \textbf{45.91} & \textbf{77.87} & 79.83 \\
\bottomrule
\end{tabular}
\end{table*}

\begin{table*}[!htbp]
\centering
\caption{Official leaderboard results of the AV2 2026 Scenario Mining Challenge on the HOTA-Temporal track.}
\label{tab:leaderboard_hota}
\begin{tabular}{clcccc}
\toprule
\textbf{Rank} & \textbf{Team} & \textbf{HOTA-Temporal($\uparrow$)} & \textbf{HOTA-Track($\uparrow$)} & \textbf{Timestamp BA($\uparrow$)} & \textbf{Log BA($\uparrow$)} \\
\midrule
1 & HYU\_OASIS & \textbf{38.50} & 52.63 & 74.32 & 77.12 \\
2 & MTL (Argonaut) & 37.04 & \textbf{55.11} & 75.50 & \textbf{80.75} \\
\textbf{3} & \textbf{MISI (AutoMine) (Ours)} & 36.38 & 49.32 & \textbf{77.21} & 76.26 \\
4 & Zelos\_Agent\_Driving & 33.45 & 45.23 & 72.45 & 73.66 \\
5 & Lilly & 32.51 & 42.06 & 71.93 & 73.11 \\
6 & DataClaw & 31.40 & 44.40 & 74.05 & 79.10 \\
\bottomrule
\end{tabular}
\end{table*}

\section{Experiments}

\subsection{Dataset and Evaluation Metrics}

We conduct experiments on the official benchmark of the CVPR 2026 Argoverse 2 Scenario Mining Challenge. The benchmark is built upon the Argoverse 2 Sensor Dataset~\cite{Wilson2023Argoverse2N}, which contains 1,000 driving logs (700 training, 150 validation, 150 test) with a total of approximately 4.2 hours of driving data. Each log lasts about 15 seconds. The sensor suite includes two 32-beam LiDARs (10 Hz), nine global shutter cameras (20 fps), HD maps, and 6-DOF ego-vehicle poses. The dataset provides 10,000 planning-centric natural language queries.
The evaluation metrics are as follows:

\noindent\textbf{HOTA-Temporal} ($\uparrow$).
The primary ranking metric, computed only on the time window where the scenario description holds. For each prompt, predictions are filtered by both class and a per-prompt confidence threshold (selected from 10 recall-based candidates), and the standard HOTA score is computed using center-distance similarity (zero-distance $=2$\,m). HOTA jointly measures detection and association quality:
\begin{equation}
    \left\{
    \begin{aligned}
        & \text{HOTA}_\alpha = \sqrt{\text{DetA}_\alpha \cdot \text{AssA}_\alpha}, \\
        & \text{HOTA} = \frac{1}{|A|}\sum_{\alpha \in A} \text{HOTA}_\alpha,
    \end{aligned}
    \right.
\end{equation}
where $A=\{0.05, 0.10, \dots, 0.95\}$ is the set of localization thresholds, and DetA, AssA denote detection and association accuracy, respectively. The final score per prompt is the maximum HOTA over the 10 candidate thresholds.

\noindent\textbf{HOTA-Track} ($\uparrow$).
Same as HOTA-Temporal, except that any track ever marked as \texttt{REFERRED\_OBJECT} in a sequence is treated as referred across all of its frames. This evaluates the full lifetime of referred tracks rather than only the description-active interval.

\noindent\textbf{Timestamp BA} ($\uparrow$).
A frame-level retrieval metric that asks whether any referred object exists in each timestamp, independent of localization. Using the optimal thresholds from HOTA-Temporal, each frame is classified as positive iff it contains at least one \texttt{REFERRED\_OBJECT}. Per prompt, we aggregate frame-level TP/FP/TN/FN and compute balanced accuracy:
\begin{equation}
    \text{Timestamp BA} = \frac{1}{2}\left(\frac{\text{TP}}{\text{TP}+\text{FN}} + \frac{\text{TN}}{\text{TN}+\text{FP}}\right).
\end{equation}

\noindent\textbf{Log BA} ($\uparrow$).
The log-level counterpart of Timestamp BA. A $(\text{log}, \text{prompt})$ sequence is positive iff \emph{any} frame contains a referred object. BA is then computed across all sequences using the same formulation as above.

\subsection{Ablation Study}

To validate the contribution of each design in AutoMine, we conduct ablation studies on the validation set, using the publicly available Le3DE2E~\cite{wang2023le3de2e} trajectories as initial input. We first compare three state-of-the-art LLMs under the single-query baseline and adopt Claude-Sonnet-4.6 as the default code generator, since it is the most stable on relation-argument ordering and fine-grained category selection. For all VLM-enhanced atomic functions, we use Qwen3.5-27B~\cite{qwen3.5} as the underlying vision-language model. We then incrementally add each component on top of this baseline. The results are shown in Table~\ref{tab:ablation}.

We observe that each component improves the system through a distinctly different mechanism rather than simply boosting all metrics uniformly. Trajectory refinement contributes a much larger HOTA-Track gain than HOTA-Temporal gain, indicating that re-linking fragmented tracklets mainly extends the lifetime of already-correct referred objects rather than enlarging the temporal window of new scenarios. Atomic function optimization brings the largest single jump, because temporally aggregated relation predicates absorb per-frame heading noise on directional queries, while VLM-enhanced functions cover attributes that are unobservable from 3D boxes alone (e.g., traffic-light state, road surface, attached cargo); the two are largely complementary and fail on disjoint subsets of queries.

Execution-driven self-refinement further improves all metrics by repairing three recurring error patterns we observe in the round-0 code: missing \texttt{reverse\_relationship} calls, over-strict thresholds that yield zero candidates, and wrong referred categories caught against \texttt{REFERRED\_DICT}. Semantic-preserving prompt augmentation yields only marginal gains on the raw baseline but becomes effective once stacked on top of the optimized pipeline, since it reduces prompt-induced variance across semantically equivalent rewrites rather than introducing new capability---an effect that is naturally amplified when the downstream pipeline is already strong.

\subsection{Leaderboard Results}

Our final submission, AutoMine, achieves excellent results on the test set of the CVPR 2026 Argoverse 2 Scenario Mining Challenge. Table~\ref{tab:leaderboard_hota} shows the leaderboard sorted by the primary metric HOTA-Temporal: we rank \textbf{3rd} with \textbf{36.38} and HOTA-Track of 49.32. Table~\ref{tab:leaderboard_timestamp} shows the leaderboard sorted by Timestamp BA: we rank \textbf{1st} with \textbf{77.21}, demonstrating the advantage of AutoMine in temporal localization accuracy.

\begin{table}[!t]
\centering
\small
\setlength{\tabcolsep}{4pt}
\caption{Official leaderboard results of the AV2 2026 Scenario Mining Challenge on the Timestamp BA track.}
\label{tab:leaderboard_timestamp}
\begin{tabular}{clcc}
\toprule
\textbf{Rank} & \textbf{Team} & \textbf{Timestamp BA($\uparrow$)} & \textbf{Log BA($\uparrow$)} \\
\midrule
\textbf{1} & \textbf{AutoMine (Ours)} & \textbf{77.21} & 76.26 \\
2 & MTL (Argonaut) & 75.50 & \textbf{80.75} \\
3 & HYU\_OASIS & 74.69 & 77.84 \\
4 & DataClaw & 74.18 & 78.67 \\
5 & EABOT\_AMD & 72.86 & 71.09 \\
6 & Lilly & 72.70 & 75.77 \\
\bottomrule
\end{tabular}
\end{table}

\section{Conclusion}
In this report, we presented \textbf{AutoMine}, a robust multimodal scenario mining framework for the AV2 2026 Scenario Mining Challenge. AutoMine addresses several key challenges in natural-language-driven scenario mining, including LLM prompt sensitivity, noisy and fragmented perception tracks, ambiguous spatial-temporal relations, and open-world visual concepts that cannot be captured by 3D trajectories alone. To this end, AutoMine refines raw trajectories, applies semantics-preserving prompt augmentation, builds robust trajectory-based and VLM-enhanced atomic functions, and further improves generated mining programs through execution-driven self-refinement using feedback from real logs. Experiments and ablation studies on the Argoverse 2 benchmark show that these components provide complementary gains, with atomic function optimization and self-refinement substantially improving actor retrieval and temporal localization. On the official leaderboard, AutoMine achieves the best Timestamp BA score of 77.21, demonstrating strong temporal localization ability, and ranks 3rd in HOTA-Temporal with a score of 36.38. These results show the promise of integrating symbolic programs, perception refinement, and visual reasoning for scenario mining.

{
    \small
    \bibliographystyle{ieeenat_fullname}
    \bibliography{main}
}

\end{document}